\documentclass{svproc}
\usepackage{amsmath,amssymb,adjustbox,makecell,xcolor,multirow,graphicx}
\usepackage{url}

\begin{document}
\mainmatter              
\title{CONVOLUTIONAL NEURAL NETWORK-BASED EFFICIENT DENSE POINT CLOUD GENERATION USING UNSIGNED DISTANCE FIELDS}
\titlerunning{Dense Point Cloud Generation}  
\author{Abol Basher \and
Jani Boutellier}
\authorrunning{A. Basher, J. Boutellier} 
%
%
\institute{School of Technology and Innovation, \\University of Vaasa,\\ Wolffintie 32, 65200 Vaasa, Finland,\\
\email{\{abol.basher,jani.boutellier\}@uwasa.fi}
}

\maketitle              

\begin{abstract}
Dense point cloud generation from a sparse or incomplete point cloud is a crucial and challenging problem in 3D computer vision and computer graphics. So far, the existing methods are either computationally too expensive, suffer from limited resolution, or both. In addition, some methods are strictly limited to watertight surfaces --- another major obstacle for a number of applications. To address these issues, we propose a lightweight Convolutional Neural Network that learns and predicts the unsigned distance field for arbitrary 3D shapes for dense point cloud generation using the recently emerged concept of implicit function learning. Experiments demonstrate that the proposed architecture outperforms the state of the art by 7.8$\times$ less model parameters, 2.4$\times$ faster inference time and up to 24.8\% improved generation quality compared to the state-of-the-art.
\end{abstract}

\section{Introduction}
\label{sec:intro}

With the advances and wider availability of real-time 3D sensors such as LiDAR and depth cameras, 3D data (e.g., point clouds) have received increased popularity in computer vision and robotics. However, the point clouds acquired from 3D sensors are often incomplete or sparsely distributed because of occlusion or sensor resolution limitations. Recovering the complete shape (structure) or generating dense point clouds is a challenging task in 3D computer vision and moreover, for many applications, it is useful to have a continuous surface representation. Although several works have focused on image-based 3D shape reconstruction, in this study, we concentrate on dense point cloud generation from a sparse (possibly incomplete) point cloud.

The conventional data types representing 3D geometry include voxels, point clouds, and meshes, each with their own limitations: voxel-based representations have a large memory footprint, which limits the output resolution; point cloud-based representations are disconnected and trivially do not allow rendering and visualization of the surface; and meshes are limited to a single topology and can have issues related to continuity. 

Recently emerged \textit{implicit representations} of 3D shapes or surfaces attempt to address these limitations. Although implicit representation based approaches \cite{Atzmon_2020_CVPR,chibane2020implicit,chibane2020ndf} are conceptually similar, they differ in inference and shape representation (occupancy values, signed and unsigned distances). The implicit function learning based approaches, which use occupancy values or signed distances as ground  truths, require closed surfaces, and can obtain the final shape (surface) in the form of a mesh through post-processing. However, in the case of occupancy values or signed distances, it is often hard to express a shape in terms of 'inside' and 'outside' (e.g.,  hollow objects or 3D scenes). Fortunately, recently \textit{unsigned distance} based implicit approaches have emerged, not suffering from the requirement of closed surfaces.

In implicit representation, a 3D shape (surface) $\mathcal{S}$ is expressed as a (zero) level set (Equation~\ref{label-set})      
\begin{align}
 \mathcal{S} = \{p \in \mathbb{R}^3 | f(p,z)=t\},
\label{label-set}
\end{align}
where $\mathrm{z}\in\mathbb{Z}\subset\mathbb{R}^{m}$ is a latent code representing a 3D shape, $p\in \mathbb{R}^{3}$ is a query point in 3D space, and $t$ is the threshold parameter (if $t=0$, then Equation~\ref{label-set} is known as zero level set). For implicit 3D shape (surface) representation, a  (convolutional) neural network $f(.)$ is trained to learn the continuous points in 3D space and predict an occupancy $f(\mathrm{p},\mathrm{z}): \mathbb{R}^{3} \times \mathbb{Z}\longrightarrow[1, 0]$, signed distance $f(\mathrm{p},\mathrm{z}): \mathbb{R}^{3} \times \mathbb{Z}\longrightarrow\mathbb{R}$, or unsigned distance $f(\mathrm{p},\mathrm{z}): \mathbb{R}^{3} \times \mathbb{Z}\longrightarrow\mathbb{R}^{+}_{0}$. 

Most recent implicit representation-based works (e.g., \cite{Atzmon_2020_CVPR}) encode a 3D shape (surface) using a single latent vector $\mathrm{z}$ and obtain a continuous presentation of the surface by learning a neural function. Due to encoding the complete shape into a single latent code, those models are prone to lose details present in the data and the original alignment \cite{chibane2020implicit} of the shape embedding in 3D space. Recently, two novel approaches have been proposed to address those limitations -- If-Nets \cite{chibane2020implicit} and neural distance fields (NDF) \cite{chibane2020ndf}. These methods encode an input shape or surface into multi-scale multi-level deep features and use these to decode the continuous shape (surface). However, NDF \cite{chibane2020ndf} relies on a convolutional neural network architecture that contains 4.6 million trainable parameters.
In this study, we propose a lightweight (0.6 million parameter) convolutional neural network architecture, \textit{LightNDF}\footnote{Source code available at: https://github.com/basher8488881/LightNDF}\cite{basher2022convolutional}, which relies on neural unsigned distance fields for implicit 3D shape representation. Compared to state-of-the-art works, LightNDF 
\begin{itemize}
\item Has 7.8$\times$ less model parameters than the next-smallest model,
\item 2.4$\times$ faster inference time than the state-of-the-art, and
\item Up to 24.3\% better generated point cloud generation quality for sparse input point clouds.
\end{itemize}

We regard the significant reduction of neural architecture size, and the reduced inference time as important factors towards wider adoption of neural implicit methods for 3D shape completion and point cloud densification. 

\begin{figure}[t]
    \centering
    \includegraphics[width=\columnwidth]{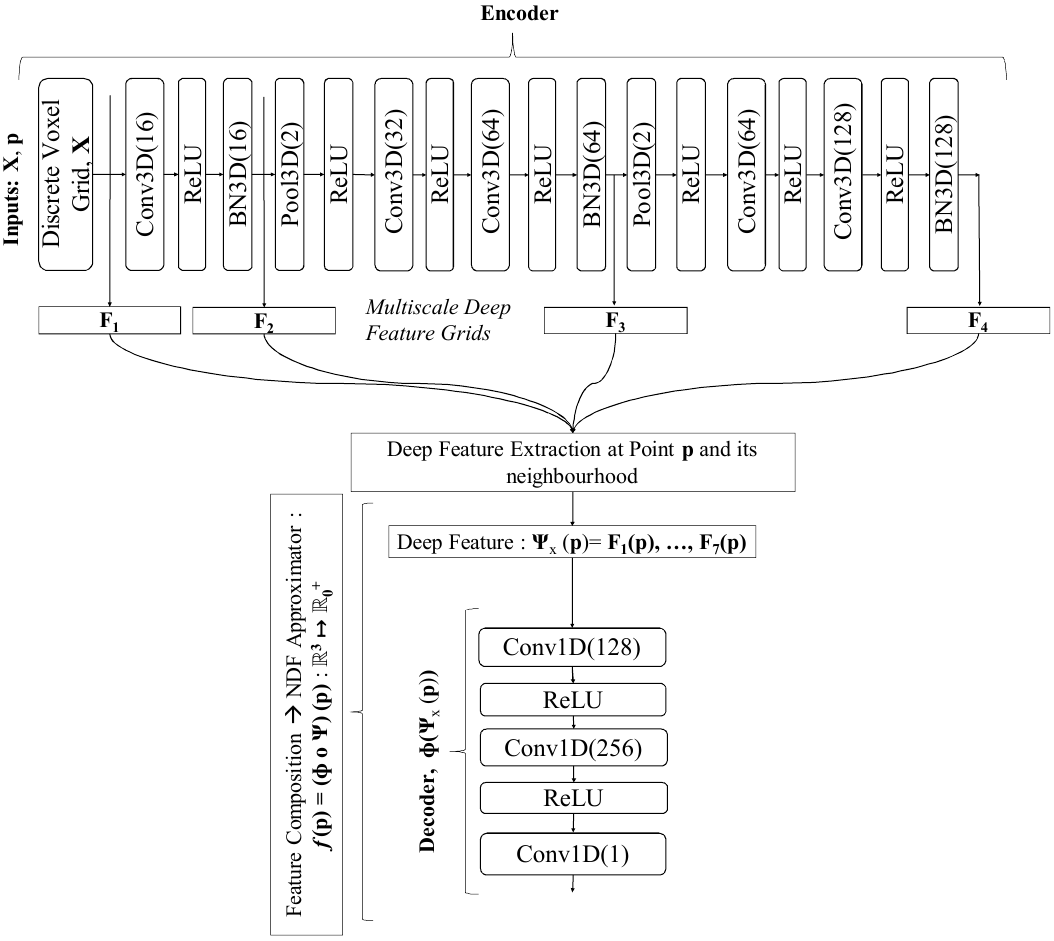}
    \caption{Proposed LightNDF architecture. Here, $\textbf{X}\in\mathcal{X}$ is input data to the encoder, while $\mathcal{X} = \mathbb{R}^{N\times N\times N}$ is a discrete voxel grid, $N$ is the input resolution, and $\textbf{p}$ is a query point.}     \label{fig:LightNDF_Architecture}
\end{figure}

\section{Related work}
\label{sec:format}

In the following, a brief review on related work in 3D representations is presented.

\subsection{Traditional representations}
Traditional 3D data representations include: (a) voxels, (b) point clouds, and (c) meshes. \textit{Voxel}-based representations are commonly used for shape (surface) learning. The convolution operation can naturally be applied to voxels, however, memory footprints increase cubically with the resolution limiting the ability of voxels to handle higher resolutions.
The \textit{point cloud} representation is another popular choice and is the output format of many sensors, e.g., LiDAR and depth cameras. The pioneering research work PointNet \cite{qi2017pointnet} has recently popularized the point cloud-based research for discriminative learning. However, due to having no connectivity information, point clouds are not well-suited for generating watertight surfaces.  
\textit{Mesh}-based representations are a more informative data type and offer connectivity information between 3D points. By deforming a template, 3D shape can be inferred using mesh-based approaches; therefore, mesh-based methods are limited to a single topological representation.  

\subsection{Implicit representation}

Recently emerged implicit function learning-based works \cite{chibane2020ndf,chibane2020implicit,Atzmon_2020_CVPR} have shown success in 3D shape representation, completion, and reconstruction. These presentations work based on zero level sets of a function (see Equation~\ref{label-set}) and learn either binary occupancy values or signed or unsigned distance functions (fields) \cite{chibane2020ndf,chibane2020implicit}. Implicit representations can represent shape and scene in a continuous fashion with various topologies. However, one important drawback of these methods, which rely on occupancy or signed distance, is that they require watertight shapes or scenes, which are not always available. Recently proposed neural unsigned distance fields (NDF) \cite{chibane2020ndf} have addressed these issues and can represent wider classes of surfaces, mathematical functions, and manifolds; unfortunately, this recently proposed approach is computationally complex. For addressing this issue, in the following we present a compact neural network architecture for unsigned distance field learning and inference.

\section{Proposed network}
\label{sec:pagestyle}

In the following, we present our lightweight convolutional neural network architecture, \textit{LightNDF} for implicit representation of the 3D geometry, inspired by the literature \cite{chibane2020implicit}. Here, we describe how the proposed LightNDF architecture learns to encode and decode a 3D shape (surface), along with the training and inference details.

\begin{figure*}[t]
    \centering
    \begin{tabular}{c c c c c}
        \includegraphics[width=0.9 in]{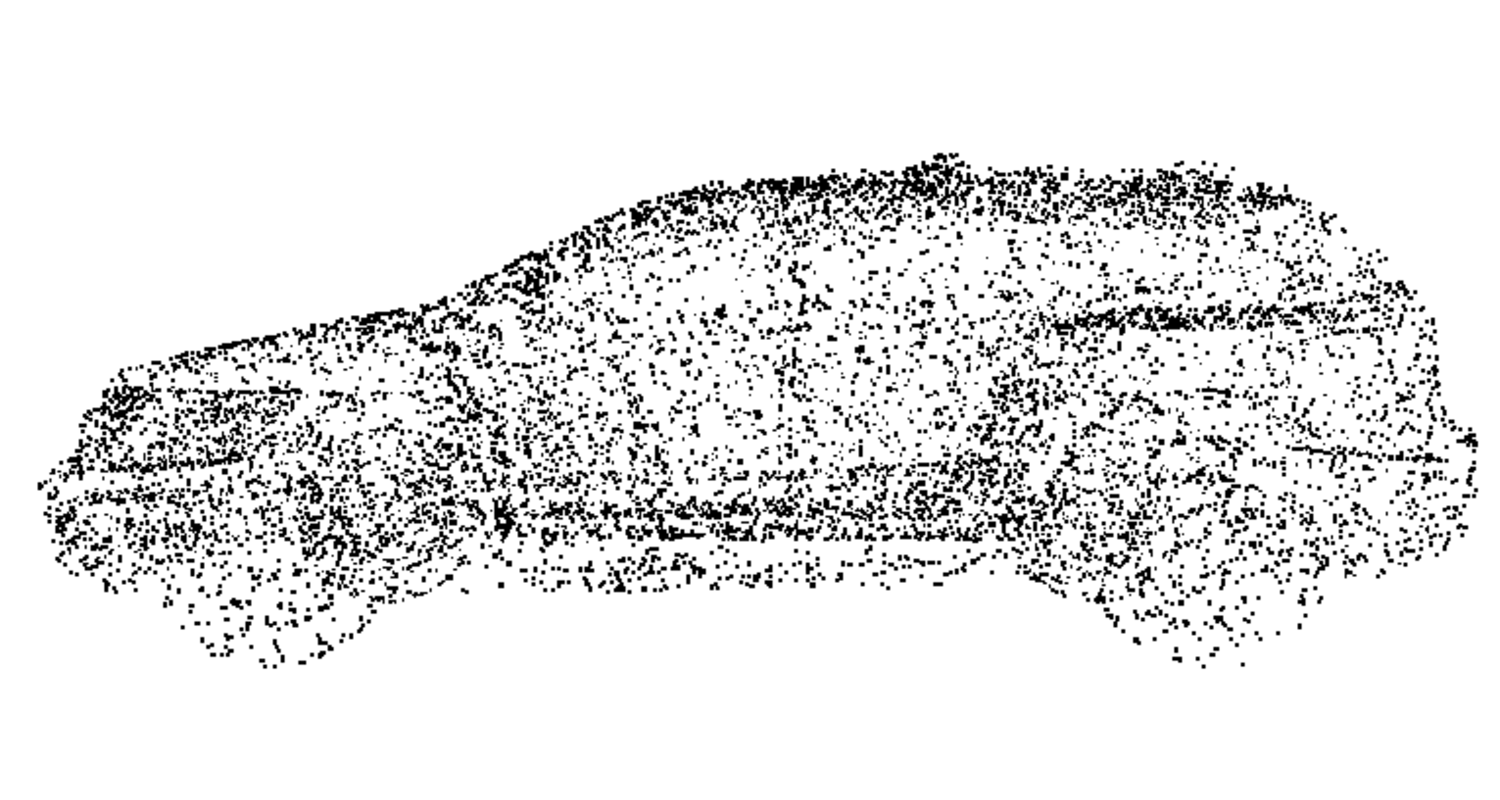} &\includegraphics[width=0.9 in]{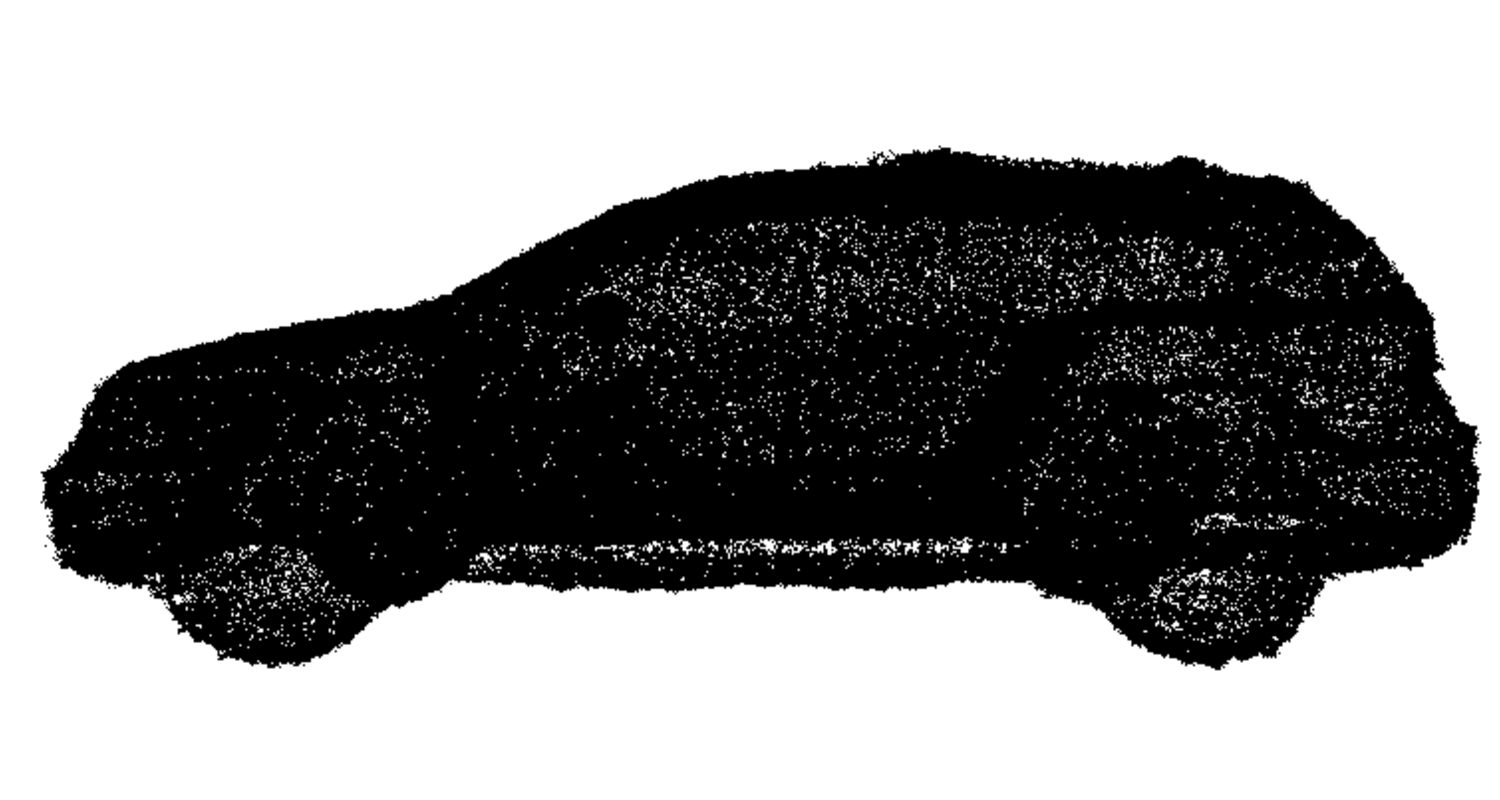}&\includegraphics[width=0.9 in]{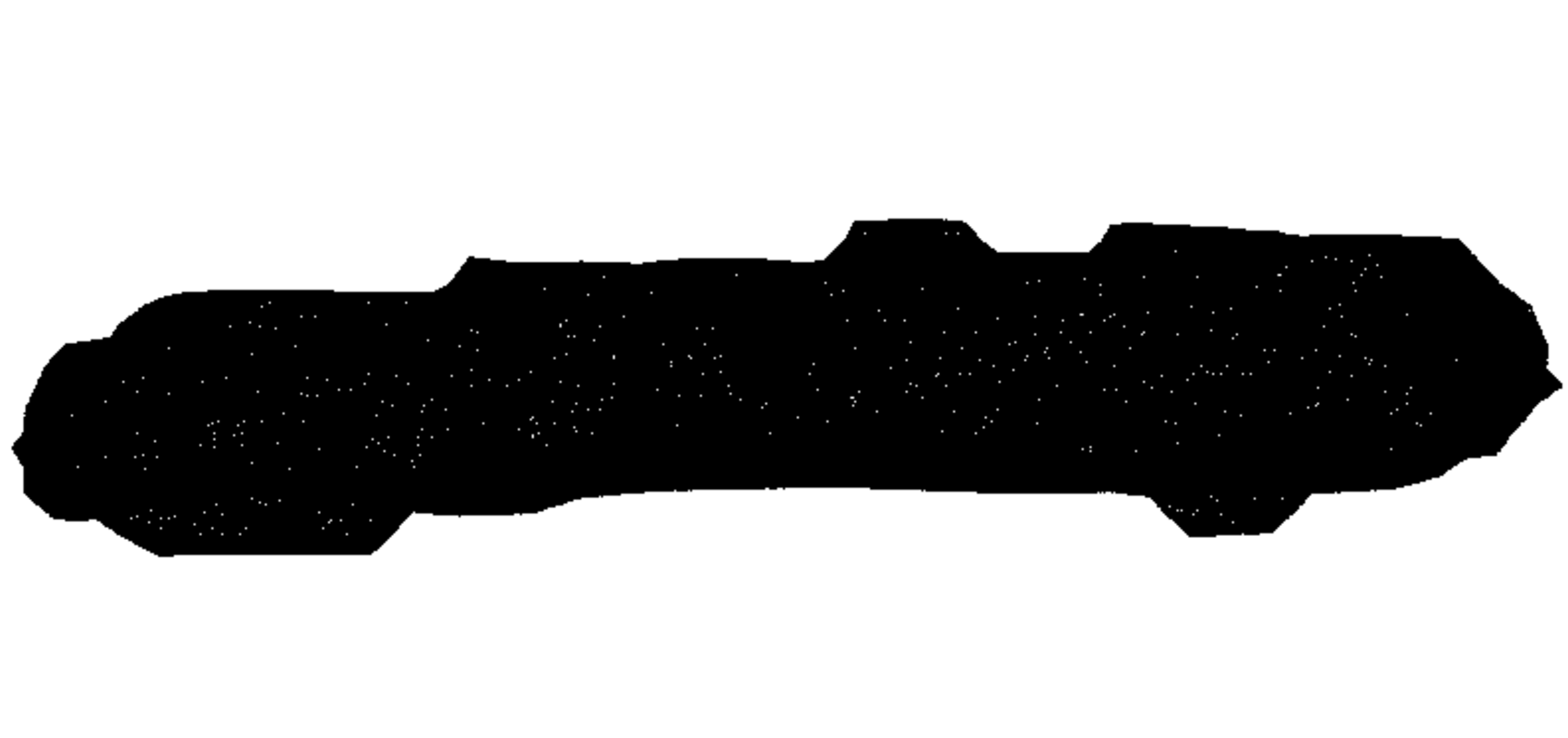}&\includegraphics[width=0.9 in]{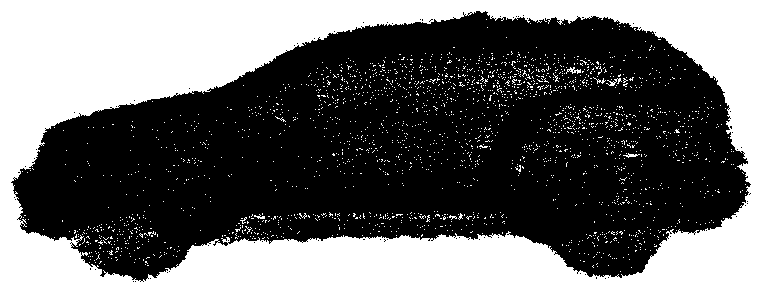}&\includegraphics[width=0.9 in]{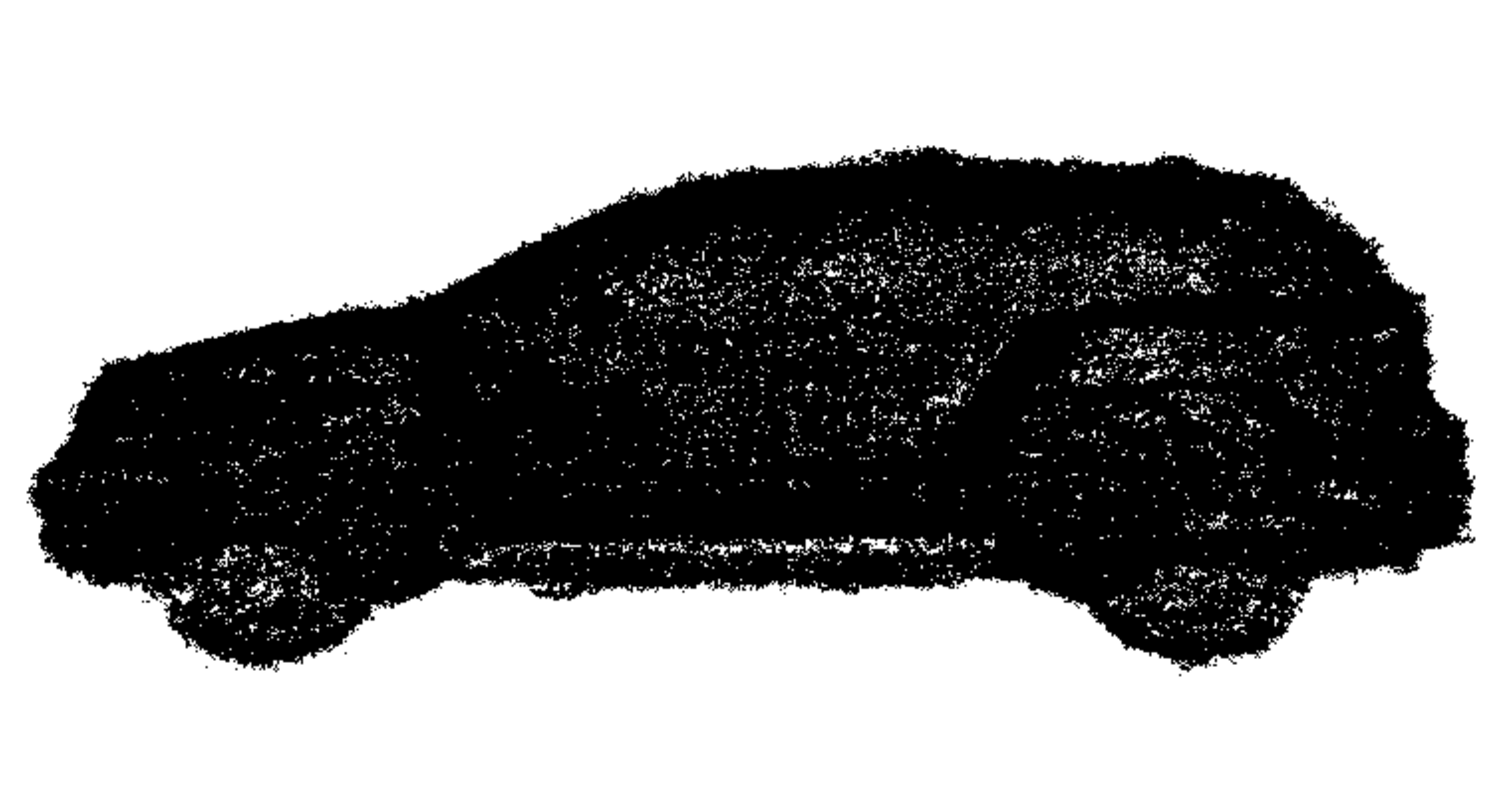}\\
        \includegraphics[width=0.9 in]{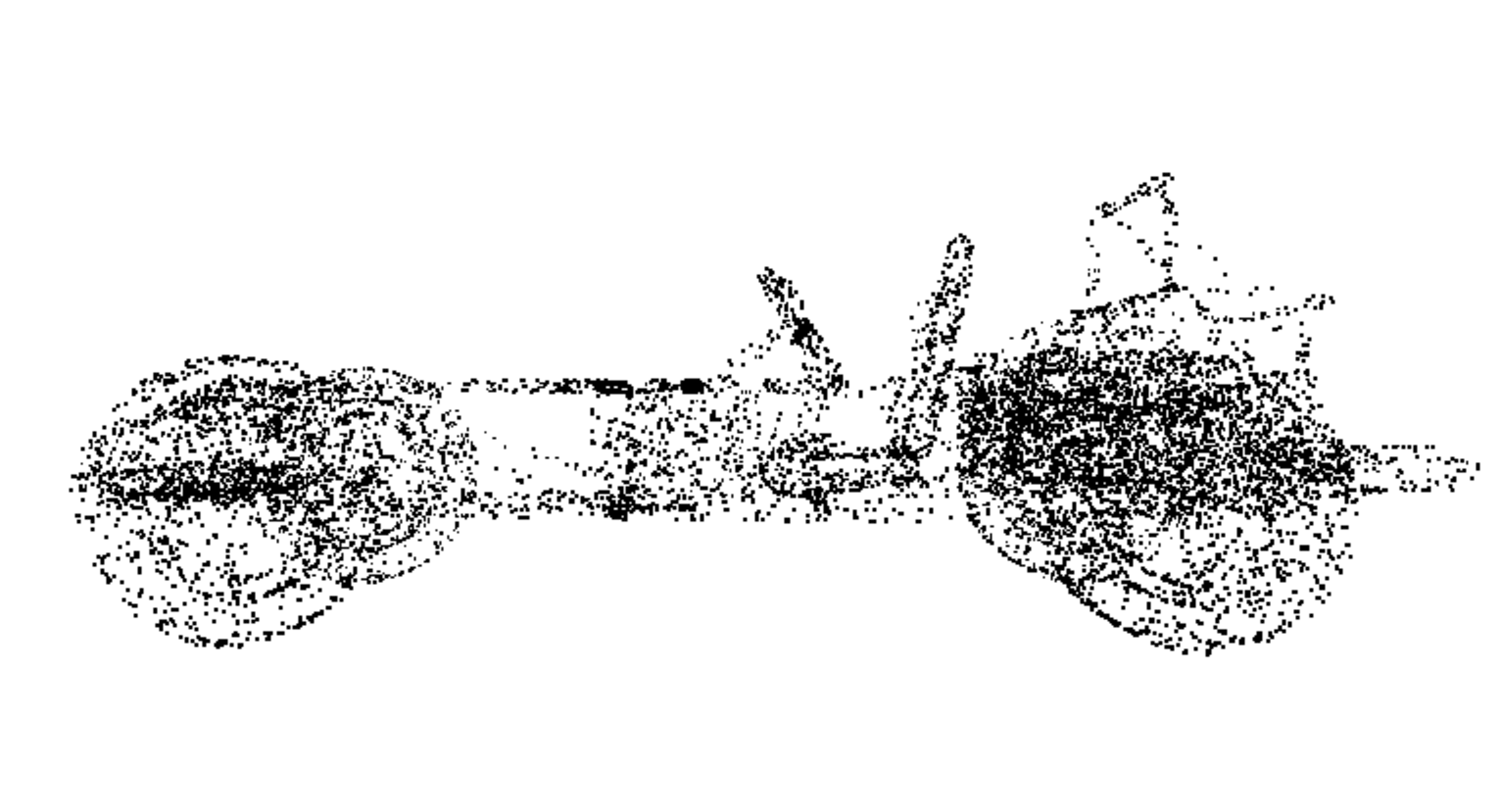}&\includegraphics[width=0.9 in]{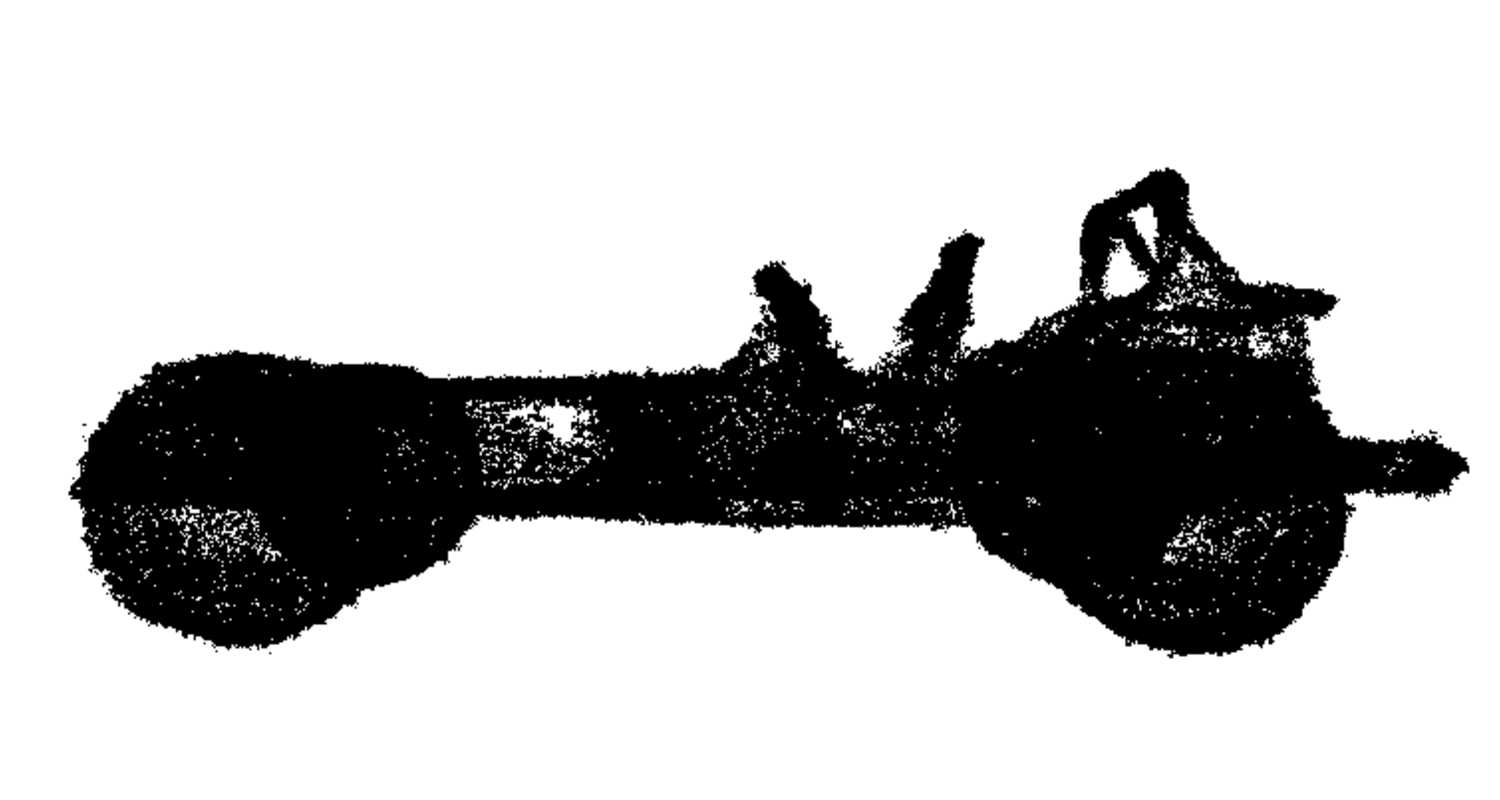}&\includegraphics[width=0.9 in]{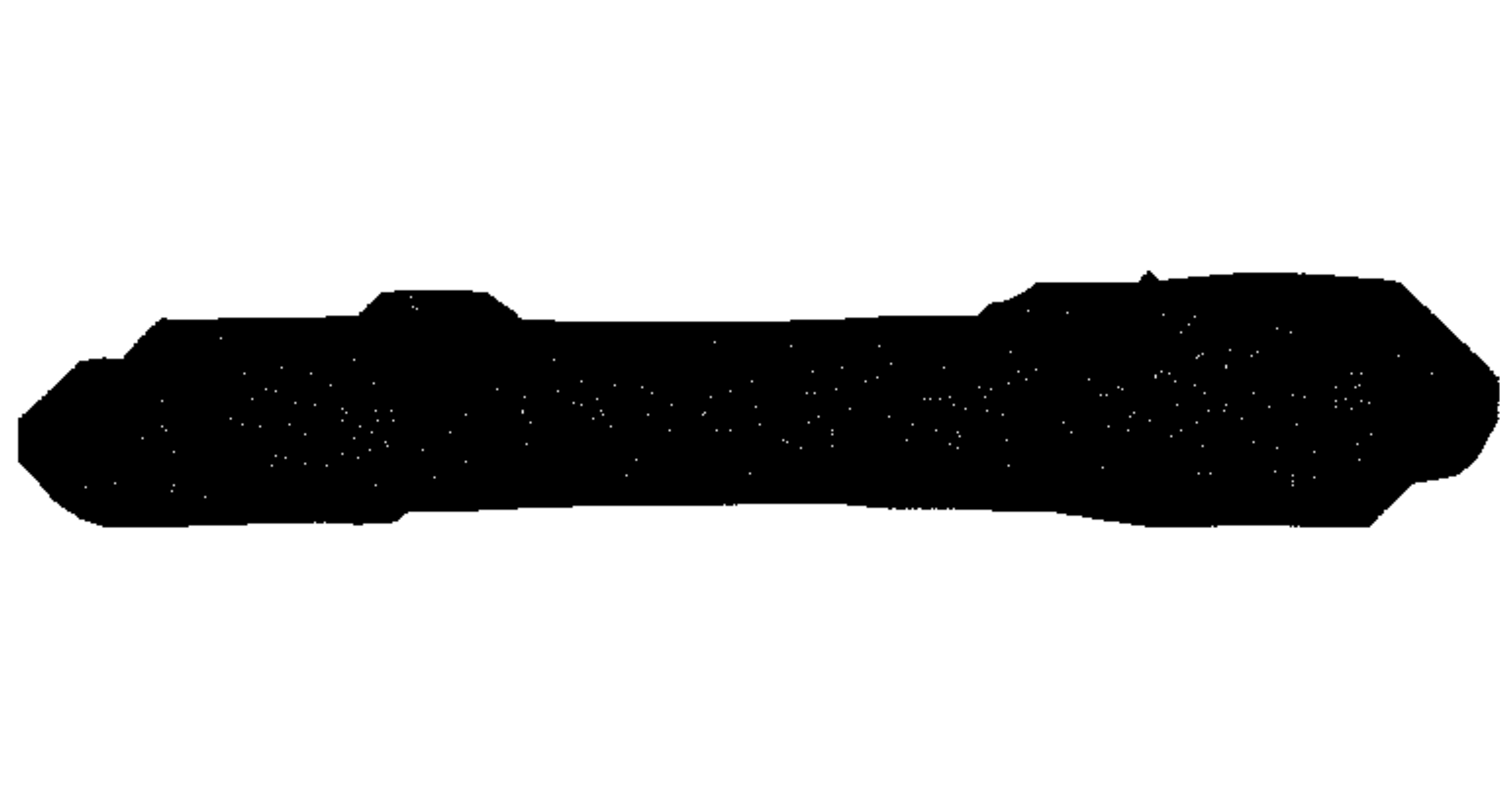}&\includegraphics[width=0.9 in]{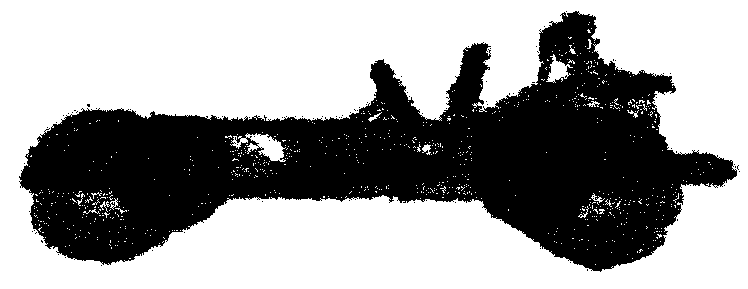}&\includegraphics[width=0.9 in]{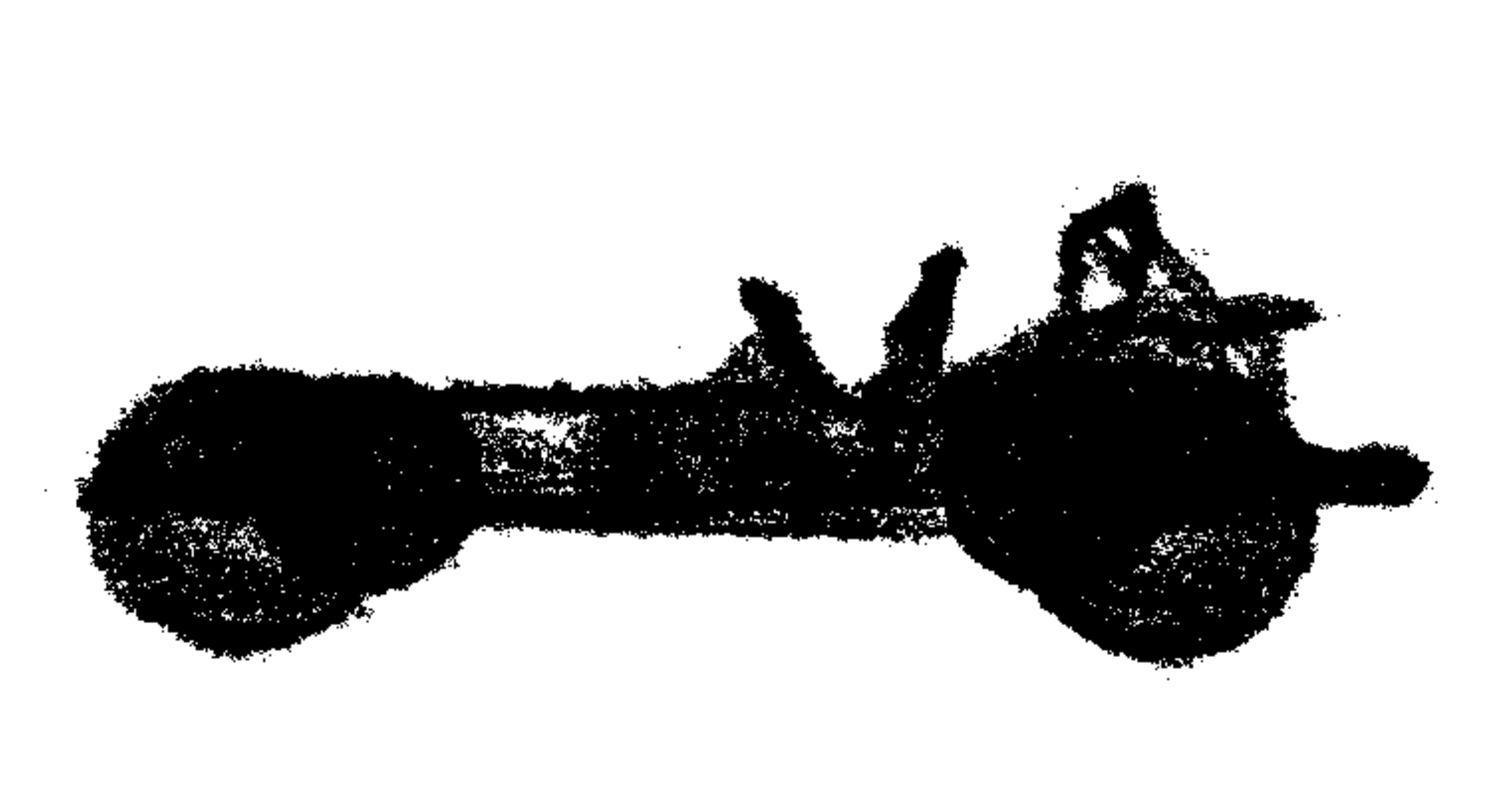} \\
        Sparse Input & NDF & SAL&GIFS&LightNDF\\ 
 
    \end{tabular}
    
    \caption{The generated dense 1M point clouds (from left to right): sparse input (10,000 points), NDF, SAL, GIFS, LightNDF (proposed). Due to the hollow input shape, SAL has significant difficulties in shape generation. Point cloud details are better visible when zoomed-in.}
    \label{fig:qualitative_results}
\end{figure*}

\subsection{Encoder}
Similar to the GIFS \cite{ye2022gifs} encoder, also our LightNDF encoder is based on a 3D convolutional neural network (CNN) that has its roots in the NDF \cite{chibane2020ndf} encoder, to handle the discrete 3D voxel grid $\mathcal{X}= \mathbb{R}^{N\times N\times N}$ generated from a sparse point cloud, $\textbf{X}\in \mathcal{X}$ of an object; however, the trainable parameter count of the LightNDF encoder is about 7.4$\times$ smaller compared to NDF. Instead of generating a single latent code of input data like other implicit representation-based approaches, LightNDF encodes the input shape using the 3D CNN into multi-scale deep feature grids $\textbf{F}_{1},...,\textbf{F}_{n},\hspace{.5em}  \textbf{F}_{k}\in\mathcal{F}^{K\times K\times K}$, with the resolution $K$ decreasing by factor of ${2^{k-1}}$ and $\mathcal{F}_{k}\in\mathbb{R}^{C}$ being a multi-channel feature where $C$ is  the number of channels. In the early stages, the extracted features $\mathcal{F}_k$ (starting at $k=1$) capture local detail of the shape, whereas the higher-level features (ending at $k=n$; in the case of LightNDF, $n=4$) capture the global structure of input data by utilizing larger receptive fields. 
Compared to other works in the field, perhaps the most similar encoder architecture can be found in one If-Net \cite{chibane2020implicit} variant that is used to extract deep feature grids for voxel-grid reconstruction (super-resolution); however, our proposed LightNDF encoder has significantly different filter configurations, and is used to extract deep feature grids with a resolution of $256^3$. This multi-scale feature extraction process enables the encoder to reason about sparse data and retain the input data details. The encoder architecture with the filter settings in each layer is shown in Fig.~\ref{fig:LightNDF_Architecture}.

 \subsection{Decoder}
 The LightNDF decoder (see Fig.~\ref{fig:LightNDF_Architecture}) consists of three 1D convolutional layers that follow the filter pattern of $( 128 \rightarrow 256 \rightarrow 1)$, with the first two layers followed by a ReLU non-linear activation function. This lightweight architectural structure makes the LightNDF decoder to have 32.7$\times$ less parameters than the NDF \cite{chibane2020ndf} decoder (60'545 vs. 1'980'673). 
 
 From the multi scale deep feature grids, a neural function,  $\Psi(\textbf{p}) = \textbf{F}_{1}(\textbf{p}), ...,\textbf{F}_{n}(\textbf{p})$, extracts deep features using trilinear interpolation at the location of a query point $\textbf{p}\in\mathbb{R}^{3}$ and its neighborhood. The extracted deep features $\textbf{F}_{1}(\textbf{p}), ...,\textbf{F}_{n}(\textbf{p})$ are fed to the decoder $\Phi( \textbf{F}_{1}(\textbf{p}), ...,\textbf{F}_{n}(\textbf{p}))$, which then predicts the unsigned distance function (UDF) to the ground truth surface, UDF(\textbf{p}, $\mathcal{S}$) = $\min_{q\in\mathcal{S}} \parallel \textbf{p} - \textbf{q}\parallel$, from the deep features, where \textbf{q} is the closest surface point from \textbf{p}. Therefore, the LightNDF or NDF approximator $f(\textbf{p})$, which approximates the points to unsigned distance, is obtained by composition and can be expressed by Equation~\ref{NDF_Approximator}: 
 \begin{align}
     f(\textbf{p}) = (\Phi \circ \Psi)(\textbf{p}):\mathbb{R}^{3}\longrightarrow\mathbb{R}_{0}^{+}.
     \label{NDF_Approximator}
 \end{align}

\subsection{Model training}
Training samples for the proposed model are generated by sampling points $\textbf{p}\in\mathbb{R}^{3}$ close to surface $\mathcal{S}$ (provided in mesh format for our dataset) and by computing the corresponding ground truth UDF(\textbf{p}, $\mathcal{S}$) using clamped regressed distance (10 cm) to learn distances. The proposed LightNDF and NDF $f(p)$ encoder and decoder were parameterized by neural weights \textbf{w} and trained jointly via mini-batch loss with the Adam optimizer and mean absolute error loss function (L1 loss); the learning rate was 0.0001. The proposed LightNDF, GIFS, and NDF models are implemented in PyTorch and were allowed to train for 45 epochs, 52 epochs, and 49 epochs (respectively) on a dual 24-GB Geforce RTX 3090 GPU. The proposed architecture was somewhat faster to train than NDF, as the duration for a LightNDF training epoch was
around 14.1 minutes, whereas for NDF a training epoch took 17.4 minutes.

\subsection{Inference and dense point cloud generation}
\label{ssec:inference}
In the inference phase, the NDF and the proposed LightNDF model generate a complete UDF surface given a sparse point cloud. The UDF surface can then be used to generate a dense point cloud, or a mesh. The regular GIFS implementation however directly generates a mesh file instead of a dense point cloud, by default. To enable comparison, the GIFS implementation was modified to use the same dense point cloud generation algorithm as LightNDF and NDF, so that it can directly the UDF surface similar to NDF and LightNDF.
We used visualization Algorithm 1 of \cite{chibane2020ndf} to extract the dense point cloud from predicted UDFs.
If the proposed LightNDF approximator $f(\textbf{p})$ perfectly estimates the true UDF(\textbf{p}), then a point \textbf{p} can be projected to the nearest surface point \textbf{q} by moving in the negative gradient direction using visualization Algorithm~1 \cite{chibane2020ndf}. Repeating the projection  \cite{chibane2020ndf} multiple times (5$\times$ in our experiments) generates millions of points on the surface. Dense point clouds can be used in many applications, such as point-based modeling, and for learning shape representations such as point cloud-based segmentation and classification \cite{qi2017pointnet}.

\section{Experimental results}
\label{sec:typestyle}

This section presents the dataset and the performance evaluation for the proposed LightNDF architecture.

\subsection{Dataset and baselines} 
The ShapeNet dataset \cite{chang2015shapenet} consists of classes of rigid objects --- for example, cars, lamps, chairs, tables, and planes, containing more than 3,000,000 models. The ShapeNetCore subset used in this work contains 51,300 unique 3D models belonging to 55 categories, where each of the models is manually verified, with alignment annotations provided. In this work, we use the Cars subset of ShapeNetCore: from the 3514 models of the Car class, about 70\% (2461) were used to train the model, 20\% (702) for testing, and 10\% (351) for validation. Using this data, the models were used to perform \textit{reconstruction of previously unseen shapes}, i.e. learning from the training set, and generating dense point clouds from the sparse samples of the test set.

The NDF neural network architecture proposed in \cite{chibane2020ndf}, the recent NDF-inspired GIFS architecture \cite{ye2022gifs} and Sign Agnostic Learning (SAL) \cite{Atzmon_2020_CVPR} were used as the baseline for evaluating LightNDF performance. We compared our LightNDF architecture against NDF, GIFS and SAL for the task of dense point cloud generation from sparse point clouds. Each architecture was trained on our own workstations using the same training and test data splits.

\subsection{Performance and comparison}

Quantitative comparison between the trained models was performed using the Chamfer-$L_{2}$ distance metric, where lower values are better, and zero means a full match between the generated point cloud and the ground truth. The results are shown for 3000 and 10000 input points in Table~\ref{tab:chamfer_L2} (in the scale of $\times{10}^{-4}$). It can be seen from the Chamfer-$L_{2}$ distance that quantitatively the proposed LightNDF model outperforms all related models (NDF, GIFS and SAL) in previously unseen shape generation for both input point cloud resolutions of 3000 points and 10000 points. All presented values are an average over the 702 test shapes.

In Fig.~\ref{fig:qualitative_results}, a visual comparison of generated 1M point dense point clouds by NDF, GIFS, SAL, and the proposed LightNDF architecture for 10000 input points are shown. Observation of the generated dense point clouds reveals no significant diffences among GIFS, NDF and LightNDF, whereas SAL shows inferior quality due to its inability to handle hollow shapes. Table~\ref{tab:times} shows the projection (inference) times for the different models, revealing that the significantly smaller LightNDF architecture is 6.7$\times$ faster than NDF, and 2.4$\times$ faster than GIFS. Inference time comparison against SAL was not meaningful, as the shape generation procedure of SAL involves significantly different shape pre-processing than the other models. 

It was observed that the number of initial sample points \textbf{p} (see Section~\ref{ssec:inference}) has a significant impact to inference time for both the NDF and the proposed LightNDF models. Inference time-optimal values of \textbf{p} were identified to be 20100 for both NDF and LightNDF; Table~\ref{tab:chamfer_L2} results are for these values of \textbf{p}. On the other hand, inference time-optimal value of \textbf{p} of GIFS is 200000. In case of \textbf{p} value of 20100, GIFS inference time increases from 1.660 and 1.663 to 2.233 s and 2.240 s in case of 10000 and 3000 input points (respectively).  The value of \textbf{p} had a negligible impact on generated point cloud quality for LightNDF, NDF and GIFS.

\begin{table}
    \centering
    \caption{Model inference time comparison for 3000 and 10000 input points. \textit{Lower values are better.\\}}
    \begin{tabular}{lcc}
    \hline
         Model&
         \makecell{Proj. time 10000}  & \makecell{Proj. time 3000}\\ 
    \hline
        NDF \cite{chibane2020ndf}&4.64 s & 4.63 s\\ 
    \hline
        SAL \cite{Atzmon_2020_CVPR}&- &-\\
    \hline
        GIFS \cite{ye2022gifs}& 1.66 s & 1.66 s\\    
    \hline
         \makecell{LightNDF (Proposed)}&0.68 s  & 0.69 s\\
    \hline
    \end{tabular}
    \label{tab:times}
\end{table}

\begin{table}
    \centering
    \caption{Model point cloud generation quality comparison. Chamfer-$L_{2}$ (CD-$L_{2}$) distances measure the accuracy and completeness of the surface between the dense point clouds (generated  by LightNDF, GIFS, SAL and NDF) and the ground truth, for 10000 and 3000 input points. CD-$L_{2}$ results are reported in the scale of $\times 10^{-4}$. \textit{Lower values are better}}
    \begin{tabular}{lccc}
    \hline
         Model&
         \makecell{\# Trainable\\ Parameters} &      \makecell{CD-$L_{2}$\\10000} 
         &\makecell{CD-$L_{2}$\\3000} \\ 
    \hline
        NDF \cite{chibane2020ndf}&4.6M&0.202&0.305\\ 
    \hline
        SAL \cite{Atzmon_2020_CVPR}&4.2M&7.444&9.170\\
    \hline
        GIFS \cite{ye2022gifs}&3.6M&0.222 &0.330\\    
    \hline
         \makecell{LightNDF (Proposed)}&0.46M&0.152&0.270\\
    \hline
    \end{tabular}
    \label{tab:chamfer_L2}
\end{table}

\section{Discussion and conclusion}
In this work, we introduced the LightNDF neural network architecture for dense point cloud generation. The proposed architecture outperforms the state-of-the-art architectures NDF and GIFS in terms of model parameter count, inference time, and quality of generated dense point clouds. We regard especially the reduction in inference time as an important development towards wider adoption of neural implicit models for 3D point cloud completion and densification. As future work we plan to study means for reducing computation time of the 3D convolutions present in the LightNDF encoder, and extend the approach towards generation of meshes.

\section*{Acknowledgements and declarations}
The authors declare that all the necessary code and data to reproduce these results is available. The authors acknowledge that this research was funded by the Academy of Finland projects CoEfNet (309903) and REPEAT (327912). The authors have no competing (financial or non-financial) interests to declare that are relevant to the content of this article. All authors whose names appear on the submission have made substantial contributions to the work.
%
%
%

\bibliographystyle{spmpsci}

\end{document}